\documentclass[preprint]{elsarticle}

\usepackage{hyperref}

\journal{Journal of \LaTeX\ Templates}

\usepackage{amssymb}
\usepackage{array}
\usepackage{geometry}
\usepackage{floatrow}
\newfloatcommand{capbtabbox}{table}[][\FBwidth]
\usepackage{ragged2e}
\justifying

\biboptions{sort&compress}

\begin{document}

\begin{frontmatter}

\title{An Ontology of Preference-Based Multi-objective Metaheuristics}

\author[add1,add2]{Longmei Li\corref{cor1}}
\ead{longmeili@nudt.edu.cn}
\author[add3]{Iryna Yevseyeva}
\ead{iryna.yevseyeva@dmu.ac.uk}
\author[add4,add5]{Vitor Basto-Fernandes}
\ead{vitor.basto.fernandes@iscte.pt}
\author[add6]{Heike Trautmann}
\ead{trautmann@wi.uni-muenster.de}
\author[add1]{Ning Jing}
\ead{ningjing@nudt.edu.cn}
\author[add2]{Michael Emmerich}
\ead{m.t.m.emmerich@liacs.leidenuniv.nl}

\cortext[cor1]{Corresponding author}
\address[add1]{School of Electronic Science and Engineering, National University of Defense Technology, Changsha, Hunan, 410073, China}
\address[add2]{Leiden Institute of Advanced Computer Science, Leiden University, 2333CA Leiden, the Netherlands}
\address[add3]{School of Computer Science and Informatics, Faculty of Technology, De Montfort University, LE1 9BH Leicester, United Kingdom\\}
\address[add4]{Instituto Universit\'ario de Lisboa (ISCTE-IUL), University Institute of Lisbon, ISTAR-IUL, Av. das For\c{c}as Armadas 1649-026 Lisboa, Portugal}
\address[add5]{School of Technology and Management, Computer Science and Communications Research Centre, Polytechnic Institute of Leiria, 2411-901 Leiria, Portugal}
\address[add6]{Department of Information Systems, University of M\"unster, 48149 M\"unster, Germany}
%
%

\begin{abstract}
User preference integration is of great importance
in multi-objective optimization, in particular in many objective
optimization. Preferences have long been considered in traditional
multicriteria decision making (MCDM) which is based on mathematical
programming. Recently, it is integrated in multi-objective metaheuristics (MOMH), resulting in focus on preferred
parts of the Pareto front instead of the whole Pareto front.
The number of publications on preference-based
multi-objective metaheuristics has increased
rapidly over the past decades. There already exist various preference handling methods and MOMH methods, which have
been combined in diverse ways. This article proposes to use
the Web Ontology Language (OWL) to model and systematize
the results developed in this field. A review of the
existing work is provided, based on which an ontology is built
and instantiated with state-of-the-art results. The OWL ontology
is made public and open to future extension. Moreover, the usage
of the ontology is exemplified for different use-cases, including
querying for methods that match an engineering application, bibliometric analysis, checking existence of combinations of preference
models and MOMH techniques,  and discovering opportunities for
new research and open research questions.
\end{abstract}

\begin{keyword}
Evolutionary computations\sep Preferences\sep Multi-objective Metaheuristics (MOMH)\sep Multicriteria decision making (MCDM) \sep OWL Ontology
\end{keyword}

\end{frontmatter}


\section{Introduction}
Most real-world optimization problems involve multiple objectives (or criteria) to be considered simultaneously. The objectives are usually conflicting,  which means improvement of one objective cannot be achieved without deteriorating other objective(s). This kind of problem is regarded as \emph{multi-objective optimization problems} (MOPs). Unlike single objective optimization resulting in a single optimum, the result of MOPs consists of multiple trade-off solutions called \emph{Pareto optimal solutions}, and the image in the objective space is referred to as \emph{Pareto front} (PF).

As \cite{SindhyaMiettinenDeb2013} indicates, conventional \emph{multiple criteria decision making} (MCDM) \cite{Miettinen2012, ViraHaimes1983}  and \emph{evolutionary multi-objective optimization} (EMO)  \cite{CoelloLamontVanVeldhuizen2007, Deb2001} are two main research fields dealing with MOPs. Here, conventional MCDM refers to methods focusing on mathematical programming, aimed at finding one solution that best fits the preferences of a decision maker (DM). In contrast, \emph{multi-objective evolutionary algorithms} (MOEAs), which belong to \emph{multi-objective metaheuristics} (MOMHs)  \cite{jones2002multi}, are population-based algorithms intended to find a set of solutions that best approximate the whole PF. Once the set is generated, it will be presented to the DM for selection of a single solution at the second step.

The two methodologies have both pros and cons as stated in \cite{DebKoeksalan2010}.  Much synergy can be gained from the collaboration of MCDM and MOMH research areas. MCDM can help to shrink the set of alternative solutions obtained by MOMH, it is also good remedy for many-objective optimization problems (MaOPs) when selection based on Pareto optimal relation is not sufficient. MOMH can assist MCDM with difficult problems which mathematical programming cannot handle, such as nondifferentiable or discontinuous functions, nonconvexity conditions, etc. These motivations are the origin of preference-based multi-objective metaheuristics (PMOMHs). The 2004 and 2006 Dagstuhl seminars witnessed an initiation of many results in PMOMHs \cite{BrankeDebMiettinenEtAl2008} when researchers in MOMH and MCDM fields gathered together to know each other and stimulate cooperation. Since then, large numbers of methods and algorithms have been proposed and published.


Good review papers exist along with the development of PMOMHs \cite{Coello2000,RachmawatiSrinivasan2006,Branke2008,bechikh2015chapter,Branke2016}. Due to the variety of approaches and contexts where preference modelling is combined, to obtain a systematical view of PMOMHs becomes increasingly complex. PMOMHs can be classified by interaction moment with the DM  \cite{PurshouseDebMansorEtAl2014} (\emph{a-priori},  \emph{a-posteriori} or  \emph{interactively}), by type of preference information used \cite{aljawawdeh2015metaheuristic} (implicit preference or explicit preference), by ways preferences are integrated in (change of objectives or change of metaheuristics), by type of the internal preference model, etc.  Overviews help researchers to figure out what has already been done, what are the relationships between different methods and what can be done in the future. Ontologies are currently the most suited way to formally describe knowledge in a standard way, by means of comprehensive notations and graphical representations, to help understand concepts and relationships in complex knowledge domains \cite{StaabStuder2013}. They have been widely investigated in knowledge management, semantic web, and electronic commerce, but underexplored for Metaheuristics and MCDM fields.

The ontology describes concepts, relationships, formal logic axioms and objects of a particular domain \cite{chaudhary2012enriching}. It allows for a shared and common understanding of the structure of information among people or software agents. It also enables reuse of domain knowledge. With the help of a PMOMH ontology, researchers can easily understand, learn and assess existing methods, discover potential combinations and seek an appropriate method for a specific problem or application. To the best of our knowledge, formal logic-based ontologies have not been used in the PMOMH  domain yet. However, in \cite{Basto-FernandesYevseyevaEmmerich2013} and \cite{KaurChaudhary2015} OWL ontologies were presented for diversity-oriented optimization and Evolutionary Computation respectively, revealing the suitability and relevance of this type of knowledge representation for the optimization algorithms research domain.

In this paper we propose a PMOMH ontology built with Prot\'eg\'e  \cite{ProtegeWebPage}, a state-of-the-art ontology editor and framework. New contributions of this paper can be summarized as follows:

(1) A comprehensive review of PMOMHs is given and the main characteristics of each algorithm are analyzed. 

(2) An OWL ontology of PMOMH is built with Prot\'eg\'e and made public for the research communities to comment, annotate, (re)use, extend and complement.

(3) Use cases of the PMOMH ontology are provided for application-algorithm matching, bibliometric analysis,  research results assessment, automatic classification and new research discovery.

(4) The PMOMH ontology provides an overview and new perspective on this scientific area. Other ontologies can be built in related scientific areas that need a systematical view and shared conceptualization. The strength of this approach relies on its generalization and knowledge inference ability.

The rest of this paper is organized as follows.  A review of state-of-the-art results in PMOMH field is given in section \ref{SectionBackground}, scientific background of the OWL ontology is introduced in section \ref{SectionOntology}. In section \ref{SectionOntologyBuilding}, the method and procedure of building the PMOMH ontology is described in detail. Use cases of the ontology are exemplified in section \ref{SectionOntologyUsecases}. Conclusions are drawn in section \ref{SectionConclusion}.

\section{Preference-based multi-objective metaheuristics}
\label{SectionBackground}

\subsection{Multi-objective Metaheuristics}
A minimization MOP is defined as follows \cite{CoelloLamontVanVeldhuizen2007} without loss of generality:
\begin{equation}
\label{equ:MOP}
\left\{ \begin{array}{l}
\min \;f(x) = {[{f_1}(x),{f_2}(x), \cdots {f_M}(x)]^T}\\
{g_j}(x) \ge 0\;\quad \;j = 1,2, \cdots ,P\\
{h_k}(x) = 0\quad {\kern 1pt} {\kern 1pt} \,\;k = 1,2, \cdots ,Q\\
x_i^L \le {x_i} \le x_i^U\;\;i = 1,2, \cdots ,n
\end{array} \right.{\kern 1pt} \;
\end{equation}
in which solution $x$ is an $n$-dimensional decision vector $x = ({x_1},{x_2}, \cdots ,{x_n}){\kern 1pt}  \in \mathbb{R} ^n$. Each variable ${x_i}$ is bounded between lower bound $x_i^L$ and upper bound $x_i^U$. ${f_i}$ is the i-th objective function and there are M objectives in total. $P$ and $Q$ specify the numbers of inequality and equality constraints, respectively. Alternatively, one might also consider combinatorial MOP in which case  $\mathbb{R} ^n$ is replaced by $\mathbb{Z} ^n$ or some other discrete search space.

The resolution of a MOP is a set of trade-off solutions, called \emph{non-dominated solutions} or \emph{Pareto optimal solutions}. The related definitions are as follows \cite{Miettinen2012}.

A vector $u = {({u_1}, \cdots ,{u_M})^T}$ is said to  \emph{dominate} another vector $v = {({v_1}, \ldots ,{v_M})^T}$, denoted as $u\; \prec \;v$, iff $\forall i \in \{ 1, \ldots ,M\} ,\,{u_i} \le {v_i}$ and $u \ne v$.

A feasible solution $x^* \in \Omega $, where $\Omega $ is the decision space of problem (1), is called a  \emph{Pareto optimal solution}, iff $\neg \exists y \in \Omega $ such that $F(y)\; \prec \;F(x^*)$. The set of all  \emph{Pareto optimal solutions} is called  \emph{Pareto set} (PS) and the image of the PS in the objective space is called the  \emph{Pareto front} (PF).

Metaheuristics are a broad family of non-deterministic optimization methods that may provide a sufficiently good solution to complex optimization problems \cite{zavala2016structural}. MOMHs have demonstrated a great success in solving MOPs, such as MOEA (which are by far the most well-known and widely used MOMH), Particle Swarm Optimization (PSO), Artificial Immune System (AIS), Ant Colony Optimization (ACO), etc. Next, we identify the characteristics of various MOMHs and the possibilities to embrace preferences.


There are three main MOEA categories, i.e., Pareto-based, indicator-based and decomposition-based \cite{ZhouQuLiEtAl2011}. Pareto-based MOEAs use Pareto dominance as selection criterion, such as NSGA-II \cite{DebPratapAgarwalEtAl2002},  SPEA2 \cite{ZitzlerLaumannsThieleEtAl2001}, and so on. Fitness assignment and diversity methods are the key elements of algorithms in this category, where preference information can be integrated in. For example, Pareto front sorting method and crowding distance calculation in NSGA-II, fitness assignment and clustering method in SPEA2 could be modified to embed preferences. Indicator-based MOEAs use an indicator to guide the search, resulting in a set of solutions that maximize the indicator. SMS-EMOA  \cite{BeumeNaujoksEmmerich2007}, HypE \cite{BaderZitzler2011}, R2-EMOA \cite{TrautmannWagnerBrockhoff2013} and POSEA \cite{yevseyeva2014portfolio} fall into this category. Computation of the indicator, e.g., Hypervolume, R2 indicator can be adjusted for the sake of preferences. Decomposition-based MOEAs transform a MOP into several single objective optimization problems. MOEA/D  \cite{ZhangLi2007} and NSGA-III \cite{DebJain2014} are two representative algorithms. Preference information can be incorporated via weight vectors and reference points for MOEA/D and NSGA-III respectively, which are core elements of the algorithms. There are also other kinds of MOEAs, e.g., memetic MOEAs combine global search with local search, differential evolution (DE) algorithms adopt new reproduction strategies based on distribution of the solutions in the current population. Coevolution algorithms evolve multiple subpopulations simultaneously to handle a complicated problem, hybrid MOEAs may import techniques from other metaheuristics such as PSO, Simulated Annealing, or algorithms from mathematical programming, machine learning.

PSO is a population-based stochastic optimization approach inspired by the behavior of bird flocking. The key idea in multi-objective PSO algorithms is how to select the global best and local best particles, which can be  incorporated with preferences. AIS is a highly parallel intelligent system, able to learn and retrieve previous knowledge to solve MOPs. Clonal Selection mechanism is an important part in AIS-based MOMHs, preferences can be combined from here. ACO was inspired by the behavior of real ant colonies and is often used for discrete optimization problems. Probability matrix, or the pheromone model could be used to link preferences. Other metaheuristics such as Simulated Annealing (SA) and Tabu Search (TS), have also been extended to tackle MOPs, could also consider preference integration.

Apart from the specialty of each algorithm, there are also common elements for preference integration. For example, changing dominance relation is one of the most widely used methods in PMOMHs.  Initialization, fitness evaluation, and termination criterion have also been used to incorporate preference information.

\subsection{Preference-based MOMH}
The ultimate goal of multi-objective optimization is to help the DM choose the most preferred solution. Hence, the integration of MCDM approaches, which facilitate such choice, becomes indispensable. Embracement of the DM's preferences into MOMHs have attracted wide attention due to the following advantages:

(1)	It is difficult for EMO to handle MaOPs (when there are more than three objectives). Since almost all of the solutions tend to be non-dominated, selection pressure will be lost when many solutions belong to the PF. Having the preference information of the DM as a selection criterion is a good way to address this problem \cite{DiPierroKhuSavic2007, KimHanKimEtAl2012}.

(2)	Inspecting and choosing solutions from the whole PF is not a trivial task for the DM. The visualization of high-dimensional space further aggravates the difficulty. If the DM gives some (even vague) information about his/her preferences, then preferred parts of the PF will be emphasized, relieving the selection burden of the DM.

(3)	Taking into account preference information, MOMHs can only focus on searching regions favored by the DM, so that the preferred solutions will be obtained more quickly. In other words, computational efforts for finding unwanted solutions will be avoided.

%
%
Due to these advantages, a large number of publications have focused on PMOMHs. There are several criteria to classify PMOMHs, such as interaction moment in \cite{PurshouseDebMansorEtAl2014}\footnote[1]{MOMHs without preferences can be regarded as \textit{a-posteriori} methods, so PMOMHs mainly focus on \textit{a-priori} and \textit{interactive} methods.} and preference information in \cite{bechikh2015chapter}. Here, we adopt the taxonomy of \cite{bechikh2015chapter} with slight changes. The categories are \emph{Reference point-based approaches}, \emph{Reference direction-based approaches}, \emph{Preference region-based approaches}, \emph{Trade-off-based approaches}, \emph{Objective comparison-based approaches}, \emph{Solution comparison-based approaches}, \emph{Outranking-based approaches}, \emph{Knee point-based approaches}.

\subsubsection{Reference point-based approaches}

Compared to other articulation of preferences, reference point\footnote[2]{It is also called aspiration level or goal vector, or preference point in \cite{AugerBaderBrockhoffEtAl2009a}. Not to be confused with reference point in Hypervolume}, first proposed by Wierzbicki \cite{Wierzbicki1980} in MCDM, is now among the most widely used methods in PMOMHs due to its natural and intuitive meaning.  In general, a reference point is a user-defined point in objective space that represents the DM's aspiration level for each objective. The optimization search can stop once a point that dominates or equal to the reference point is found. When the reference point is infeasible, the closer a solution is to the reference point, the more it is preferred.

The earliest PMOMH is regarded to be multi-objective genetic algorithm (MOGA) \cite{FonsecaFlemingothers1993} proposed by Fonseca and Fleming. A goal vector was used to define a new dominance relation that gives higher priority to objectives that do not satisfy the goals. This method was extended by Tan et al.  \cite{TanKhorLeeEtAl2003} to incorporate hard/soft priority information and capable of dealing with multiple reference points. Utilizing the reference point to change dominance relation was also adopted in g-dominance \cite{MolinaSantanaHernandez-DiazEtAl2009} and r-dominance \cite{BenSaidBechikhGhedira2010}. In g-dominance, solutions that satisfy all the aspiration levels or fulfill none of them are preferred over solutions that satisfy some of the aspiration levels. In r-dominance, solution $x$ is referred to r-dominate solution $y$ when $x$ Pareto dominates $y$, or the weighted Euclidean distance between $x$ and the reference point is smaller than the distance between $y$ and the reference point to a predefined extent. g-dominance does not comply with the Pareto dominance, which means a dominated solution may be preferred to the solution that dominates it. Fortunately, r-dominance preserves the Pareto dominance.  Tchebycheff preference relation \cite{JaimesMontanoCoello2011} is another dominance relation changed by reference point.  A Region Of Interest (ROI) is defined with reference point and a threshold, solutions in this ROI are compared by the classical Pareto dominance relation, while solutions outside of the ROI are compared by their Tchebycheff achievement function values.

Apart from dominance relation, reference point can also be used to change crowding distance, as it is done in R-NSGA-II \cite{DebSundar2006}. Here, the original crowding distance is replaced by weighted Euclidean distance to the reference point. R-NSGA-II was later extended by Siegmund et al. \cite{siegmund2012finding} to speed up the search under a limited number of evaluations, and was also extended by Filatovas et al.  \cite{filatovas2015synchronous} to consider several scalarizing functions simultaneously. In \cite{Fei-yueYu-shiLi-pingEtAl2012}, a positive reference point and a negative reference point are integrated in crowding distance to form a new metric called \emph{similarity}. The aim is to find Pareto optimal solutions that are close to the positive reference point and far away from the negative reference point.

Weighting achievement scalarizing function genetic algorithm (WASF-GA) \cite{RuizSaboridoLuque2015} applies an achievement scalarizing function (ASF) to classify the individuals into several fronts. The main purpose of this method is to generate a well-distributed set of non-dominated solutions approximating the ROI defined by a reference point. The authors also published the interactive version of WASF-GA \cite{RuizLuqueMiettinenEtAl2015}.
MOEA/D \cite{ZhangLi2007} relies on a decomposition strategy such as weighted-sum or Tchebycheff approach to convert a multi-objective problem into single-objective problems. Reference point has also been imported to change weight vectors in \cite{mohammadi2012reference,MohammadiOmidvarLiEtAl2014}, for finding preferred solutions. NSGA-III \cite{DebJain2014} utilizes a number of well-spread reference points for MaOPs, it can also focus on preferred part of the PF using user-supplied reference points. Under such conditions, NSGA-III belongs to PMOMH.

Reference points have also been applied to indicator-based MOEAs. In PBEA \cite{ThieleMiettinenKorhonenEtAl2009}, $\varepsilon $-indicator has been modified to incorporate ASF values. Aspiration Set EMOA \cite{Trautmann2014} considers a set of reference points to guide the search, with averaged Hausdorff distance as quality indicator. Preferences can be combined with R2-indicator by position of reference point, restriction of weight space and density of weight distribution \cite{WagnerTrautmannBrockhoff2013}. R2-EMOA \cite{TrautmannWagnerBrockhoff2013}  aims at solutions maximizing the preference-based R2-indicator.

Reference point-based PSO using a Steady-State approach (RPSO-SS) \cite{AllmendingerLiBranke2008} proposes to use distance to reference points as selection  criterion, for the sake of finding a set of solutions near the reference points.  Multi-objective Differential Evolution and PSO (MDEPSO) \cite{WickramasingheLi2009} hybridizes two MCDM methods: reference point and light beam search \cite{WickramasingheLi2009}. In  \cite{chica2015interactive},  g-dominance \cite{MolinaSantanaHernandez-DiazEtAl2009} is combined with memetic multi-objective ant colony optimization  for solving a real industrial problem of assembly line balancing.


\subsubsection{Reference direction-based approaches}

Reference direction method \cite{KorhonenLaakso1986} and light beam search \cite{JaszkiewiczSlowinski1999}, which can be considered as extensions of reference point method in MCDM, are also integrated in MOMH. LBS-NSGA-II  \cite{DebKumar2007a} and RD-NSGA-II \cite{DebKumar2007} are the representative algorithms.  Light beam search has also been considered in MOEA/D \cite{ma2016moea}, multi-objective Differential Evolution and PSO (MDEPSO) \cite{WickramasingheLi2009} and multi-objective immune algorithm \cite{liu2013preference}. Branke and Deb proposed a biased crowding distance approach \cite{BrankeDeb2005} to find a biased distribution on the PF. A direction vector should be provided by the DM to define an iso-utility function, the crowding distance in NSGA-II is modified to focus solutions parallel to this iso-utility function. This approach was regarded as an objective scaling method in \cite{Branke2008}.

\subsubsection{Preference region-based approaches}
Instead of using a single point to represent preferences, a preference region in the objective space favored by the DM is also popular. Usually, it is difficult for the DM to give a precise reference point. Hence, a vague range of candidate values for such point could be used as a reasonable alternative. Moreover, functions that reflect the objective values and DM's degree of satisfaction are included in this category, two representatives are Desirability Function and density function in the objective space\footnote[1]{We call it density  function here, although originally it is referred to as weight function. As a weight we would rather consider an integral over the density function.}, as discussed next.

Desirability functions (DFs) \cite{derringer1980simultaneous,Harrington1965} are widely investigated for its simple and intuitive meaning. DFs nonlinearly transform the objective values into the desirability domain [0, 1]. 0 stands for unacceptable and 1 represents fully satisfied.  By changing the values of objectives corresponding to 0 and 1, the DF can focus the search on different regions of the PF. DFs have already been successfully introduced in combination with NSGA-II \cite{TrautmannMehnen2009}, MOPSO \cite{MostaghimTrautmannMersmann2010} and SMS-EMOA \cite{WagnerTrautmann2010} on both benchmark problems and practical problems. In order to alleviate the computational burden of SMS-EMOA, Trautmann et al. proposed to use the desirability index (DI), which is a DF-based scalarization, as the second-level selection criterion in the non-dominated sorting  \cite{TrautmannWagnerBiermannEtAl2013}. 

Another popular method to integrate preference is a density function in the objective space. In  \cite{ZitzlerBrockhoffThiele2007, AugerBaderBrockhoffEtAl2009a,BrockhoffBaderThieleEtAl2013}, weighted hypervolume is used to guide the search towards ROI by utilizing a variety of density functions in the objective space, including stressing one objective, one  reference point, a rectangle region and so on. This kind of preferences can also be integrated with NSGA-II and SPEA2, as demonstrated in \cite{FriedrichKroegerNeumann2011}. It can yield similar results compared to the hypervolume approach and requires less computational effort.  Desirability Functions and density functions can be integrated. Emmerich et al. use Desirability Function to define a density in the objective space and provide a probabilistic interpretation for it \cite{EmmerichDeutzYevseyeva2014}.

Karahan and K\"oksalan proposed a territory defining steady-state elitist evolutionary algorithm (TDEA)   \cite{KarahanKoeksalan2010} which defines a territory around each individual solution to prevent crowding. They also introduced a preference-based approach, called prTDEA, to assign different sizes of territories for preferred regions and non-preferred regions. Preferred regions have smaller territories so that a denser coverage can be achieved around them. An interactive version of this method was proposed in \cite{KoeksalanKarahan2010}. A hyperplane in the objective space is constructed to articulate preferences in   \cite{NarukawaSetoguchiTanigakiEtAl2015}. The DM can visually specify his/her preferences by center and spread vectors of Gaussian functions on the hyperplane. This preference model is embedded into the framework of NSGA-II and applied to many-objective knapsack problems \cite{TanigakiNarukawaNojimaEtAl2014}. A reference vector-guided EA (RVEA) for many-objective optimization was proposed recently \cite{ChengJinOlhoferEtAl2016}. It can not only find the whole PF as most \textit{a-posteriori} methods do, but also target a preferred subset inside a ROI (which is defined by a central vector and a radius). In interactive preference-inspired co-evolutionary algorithm (iPICEA-g)  \cite{WangPurshouseFleming2013}, candidate solutions and goal vectors are co-evolved to focus on a ROI, which is defined by user-provided reference point, weight and search range. It also allows DM to brush ROI in the objective space, alleviating the burden of setting parameters. Yang et al. proposed a method for effectively approximating a preferred part of PF based on multi-objective efficient global optimization (EGO) \cite{yang2016preference}. It uses truncated expected hypervolume improvement (TEHVI) as infill criterion, making it possible to focus the search within user-supplied region.  A hybrid multi-objective immune algorithm (HMIA) was devised with a new concept of dominance, called region-dominance \cite{jiao2010hybrid}. It enables DM to obtain an efficient set of solutions in his/her preferred region without using any scalarizing function.


\subsubsection{Trade-off-based approaches}
Trade-offs have been widely investigated in MCDM literature. According to Miettinen et al. \cite{MiettinenRuizWierzbicki2008},  trade-offs can be objective or subjective. On the one hand, a trade-off that depends on the structure of the problem ( i.e., the change in one objective with respect to change of another one, when moving from a feasible solution to another), is objective trade-off.  On the other hand, a trade-off that represents how much the DM is ready to sacrifice the value of some objectives in order to improve another objective(s), is called subjective trade-off. PMOMHs often deal with subjective trade-offs.

In the guided MOEA proposed by Branke et al. \cite{BrankeKauslerSchmeck2001}, subjective trade-offs are given by the DM in the form of statement: ``one unit improvement in objective $i$ is worth at most ${a_{ji}}$ units degradation in objective $j$ " . The basic idea of this approach is to modify the dominance relation: a solution $x$ is  preferred to a non-dominated solution $y$ if it does not violate the specified subjective trade-offs.

Apart from dominance relation, trade-offs can also be employed to sort additional fronts in the best non-dominated set, just as pNSGA-II \cite{ShuklaHirschSchmeck2010} did. In this method a set ${F_{0}}$ is found which satisfies the acceptable trade-offs from the current best front ${F_{1}}$. Solutions in this front are assigned with rank 0, which is better than the rank of ${F_{1}}$.  The remaining steps of pNSGA-II are the same as of  NSGA-II.

Trade-offs can also be integrated with set quality indicators. A concept of cone-based hypervolume indicators (CHI) is proposed and theoretically investigated in \cite{EmmerichDeutzKruisselbrinkEtAl2013, ShuklaEmmerichDeutz2013}.  The idea is to use a family of polyhedral cones with scalable opening angle $\gamma $ to express preferences in the sense of trade-off constraints.  Furthermore, the authors presented two searching algorithms to obtain solutions that are compatible with the given preference model, i.e., to find a subset of solutions that maximize the CHI.

\subsubsection{Objective comparison-based approaches}
Objective comparison is referred to as statements of preference on objectives, such as ``prefer ${f_{1}}$ to ${f_{2}}$" or classify objectives qualitatively as ``most important, important, less important", or describe their importances quantitatively as weights.

Early researches focused on weights  as preference articulation \cite{CvetkovicParmee2002}, but sometimes it is  difficult for the DM to specify weights accurately. Vague values, or fuzzy preference became popular. Jin and Sendhoff  transformed fuzzy preferences on objectives into weight intervals \cite{JinSendhoff2002}, this method is an extension of method proposed in \cite{cvetkovic1999use}.
Rachmawati and Srinivasan introduced relative importance of objectives, including strict preference, equality of importance, and  incomparability between pairs of objectives \cite{RachmawatiSrinivasan2010}. An elicitation algorithm was also proposed to assist a human DM to construct a coherent overall preference model, which is then combined with NSGA-II to obtain a subset of PF. Brockhoff et al. utilized two preference articulation approaches in the interactive W-HypE \cite{BrockhoffHamadiKaci2014}. One is to let the DM choose the most preferred solution from a set of alternative solutions (this approach belongs to ``Solution comparison-based approaches", see the next subsection). The other approach uses comparative preference statements, such as ``prefer ${f_1} < 0.5$  to  ${f_2} < 0.3$" \cite{BrockhoffHamadiKaci2014}.  This information is then transformed to parameters in the objective space density function used by W-HypE \cite{BrockhoffBaderThieleEtAl2013}. Preference information was provided by an objective pairwise comparison matrix and combined with fuzzy measure and fuzzy integral in \cite{LeeKim2011}. This preference is coupled with  multi-objective particle swarm optimization  \cite{LeeKim2011} and multi-objective quantum-inspired evolutionary algorithm \cite{KimHanKimEtAl2012}. 

\subsubsection{Solution comparison-based approaches}
Solution comparison is often used in interactive MOMHs where the DM does pairwise comparison, rank or grade a sample of representative solutions, or choose the best (and/or worst) solution(s) in a sample set. This information will guide the search to preferred part of the PF.

Phelps and K\"oksalan proposed an interactive evolutionary metaheuristic for multi-objective combinatorial optimization \cite{PhelpsKoeksalan2003}. They assumed a linear value function and used the DM's pairwise comparisons to determine the most discriminating function compatible with the preferences. A more flexible algorithm \cite{koksalan2007evolutionary} was introduced by them for all cases between the two extremes of perfect information and no information. If the utility function of the DM is known precisely, the algorithm will result in the optimum for that function; if no preference information is given, the algorithm will find the whole PF.

Fowler et al. adopted a  general quasi-concave utility function to represent the DM preferences   \cite{FowlerGelKoeksalanEtAl2010}. The best and worst solutions in the sample set are selected by the DM and used to create convex preference cones in the objective space, which are utilized to identify inferior solutions. The Necessary-preference-enhanced Evolutionary multi-objective Optimizer (NEMO-I) \cite{BrankeGrecoSlowinskiEtAl2010} presented by Branke et al. is a combination of interactive evolutionary multi-objective optimization with robust ordinal regression (ROR). At regular interactions, the DM is asked to compare some pairs of solutions in the current population. The whole set of additive value functions compatible with this preference information is utilized and integrated in a modified version of  NSGA-II. To reduce the computational complexity, which is a main shortcoming of  NEMO-I, the authors proposed NEMO-II  \cite{BrankeCorrenteGrecoEtAl2016} which compares each solution to all other solutions as a set, instead of pairwise comparison. While NEMO-I and NEMO-II both consider whole sets of value functions compatible with the preference information, NEMO-0 \cite{BrankeGrecoSlowinskiEtAl2015} uses only the most representative value function.  Similar preference model is used in PI-EMO-VF \cite{DebSinhaKorhonenEtAl2010}. Once a most discriminating value function has been identified, it is combined in a new domination principle as well as a preference-based termination criterion. Sinha et al. proposed interactive EMO algorithm based on polyhedral cone (PI-EMO-PC)  \cite{SinhaKorhonenWalleniusEtAl2010}, in which the cone is constructed according to DM's best selection in a set of alternative solutions, and it is used to eliminate a part of the search space for a more focused search.  The construction and application of the  polyhedral cone were extended for interval MOPs in  \cite{gong2013evolutionary}.

Dominance-based Rough Set Approach (DRSA) is an MCDM approach based on ``$if..., then...$" rules deduction  
 \cite{GrecoMatarazzoSlowinski2008}. It has been integrated in interactive EMO \cite{GrecoMatarazzoSlowinski2010}, where two schemes were proposed to collect user preference. One is to make the DM sort some solutions into ``relatively good" and ``others", the other requires pairwise comparison among representative solutions. Then, a set of decision rules is induced and used for ranking solutions in the current population.

Some researchers use machine learning approaches to learn the value function, such as support vector machine (SVM) \cite{BattitiPasserini2010}, Artificial Neural Networks (ANN) \cite{todd1999directed, pedro2013decision,pedro2014inspm}, instance-based supervised online learning \cite{krettek2009interactive}.

Cruz-Reyes et al. introduced the Hybrid-MultiCriteria Sorting Genetic Algorithm (H-MCSGA) \cite{Cruz-ReyesFernandezGomezEtAl2014}, in which the selective pressure based on dominance is strengthened by assigning solutions into ordered categories. A reference set of solutions belonging to different  categories is kept and updated to capture the preferences. The goal of this method is to find non-dominated solutions belonging to the best category.

Solution comparison can also be incorporated with Pareto memetic algorithm \cite{Jaszkiewicz2007}, MOEA/D \cite{GongLiuZhangEtAl2011} and Indicator-based MOEA \cite{zitzler2010set,chugh2015interactive}.

\subsubsection{Outranking-based approaches}
An outranking relation is a binary relation $S$ defined on the set of potential solutions (also called actions) $A$ such that $aSb$ if there are enough arguments to decide that $a$ is at least as good as $b$, whereas there is no essential argument to refuse that statement \cite{FigueiraGrecoEhrgott2005}. Outranking relation is widely investigated in the MCDM field.

Fernandez et al. proposed a Non-outranking Sorting Genetic Algorithm (NOSGA) \cite{FernandezLopezBernalEtAl2010} to consider a binary fuzzy preference relation that expresses the degree of truth of predicate ``$x$ is at least as good as $y$". Pareto dominance is replaced by outranking relation and it searches for the non-strictly outranked frontier which is a subset of the PF. The method is extended to increase the selective pressure towards the best compromise solution later in NOSGA-II \cite{FernandezLopezLopezEtAl2011}. A hybrid evolutionary simulated annealing (HESA) \cite{oliveira2013hybrid} was designed to improve Evolutionary Algorithm Based on an Outranking Relation (EvABOR) \cite{oliveira2013comparative}. The DM's preferences are elicited and exploited using the principles of the outranking-based ELECTRE TRI method.

\subsubsection{Knee point-based approaches}
Knee points of PF are characterized by the fact that a small improvement in one objective will arouse a large deterioration in other objectives. It is similar to trade-off-based approaches, but in trade-off-based approaches the trade-off constraint is explicitly provided by the DM, while in knee point-based approaches, no explicit preference is given, knee point (point with the maximal trade-off) is regarded as the most preferred by the DM.

Bechikh et al. proposed KR-NSGA-II \cite{BechikhBenSaidGhedira2010} based on R-NSGA-II \cite{DebSundar2006}, in which mobile reference points are used to modify crowding distance in the traditional NSGA-II. In each generation, knee points in the current population are found and serve as reference points, the method can be used  both a-priori and interactively. TKR-NSGA-II\cite{BechikhSaidGhedira2011} is an enhanced version of  KR-NSGA-II\cite{BechikhBenSaidGhedira2010} where the approach to find knee points has been improved. 


\section{Introduction to Ontologies}
\label{SectionOntology}
Ontologies are content theories (i.e., theories that explain the specific factors that motivate behavior) about the sorts of objects, properties of objects, and relations between objects that are possible in a specified domain of knowledge \cite{chandrasekaran1999ontologies}. First addressed by the Knowledge Acquisition Community and aimed at ``knowledge modeling", ontology design and engineering is now an important research and application topic in many domains, such as the Semantic Web, knowledge management, e-Learning, e-Commerce, etc. The most widely accepted definition of ontology in this context is given by Gruber \cite{Gruber1993}: ``An ontology is a formal explicit specification of a shared conceptualization for a domain of interest." It is formal logic-based, allowing for mathematical treatment and computational processing. It has explicit specification, because concepts, properties, functions and axioms are explicitly defined. It is shared with standard annotations, aiming at representation of consensual knowledge. In essence, it is conceptualization, or abstract model of some  phenomena in the world.

In February 2004, the World Wide Web Consortium (W3C) announced the final approval of two key Semantic Web technologies. One of them is the Web Ontology Language (OWL) \cite{HorridgeJuppMoultonEtAl2009}. It makes use of the eXtensible Markup Language (XML) for the definition of text-based documents syntax/structure, by means of tags that can be added to parts of the text documents, promoting interoperability between applications that exchange machine-comprehensible information.


OWL-DL (Description Logics) is a widely used sub-language of OWL. One of its key features is the capability of being processed by a reasoner. 
An OWL-DL ontology was built for the PMOMH knowledge domain and will be presented in the next sections of this paper.


\emph{ Prot\'eg\'e} is a free, open-source platform, which serves a growing number of users to construct domain models and knowledge-based applications with ontologies \cite{ProtegeWebPage}. 
It was developed and maintained by Stanford Center for Biomedical Informatics Research (BMIR) and now has more than 300 thousand registered users. People from different backgrounds can publish and import OWL ontologies for research freely.

According to Noy \cite{NoyMcGuinnessothers2001}, the reasons to develop an ontology are given below:

(1) \textit{To share common understanding of the structure of information among people or software agents.} It is of high value and interest in a knowledge domain to share and use the same underlying ontology. Additionally, computer agents can  extract and aggregate information from different sources, answer more complex user (or other computer systems) queries, as well as providing or using as input each others' knowledge basis. As PMOMH connects MCDM and MOMH communities, there are plenty of relevant shared concepts and relations between concepts in these domains. Hence, ontology can help to define a machine-interpretable vocabulary in this domain, based on which reasoning and further analysis can be done.

(2) \textit{To enable reuse of domain knowledge}. This is one of the main benefits of developing an ontology. If one group of researchers develops an ontology, others can reuse, compose and extend it for their domains. For example, PMOMH domain includes concepts such as preference information (including reference points, reference direction, etc.), MOPs and MOMHs. Preference information can be reused in MCDM community and detailed concepts in MOMHs  or MOEAs can also be of interest in research domains of (single objective) metaheuristics or evolutionary algorithms.

(3) \textit{To make domain assumptions explicit}. Explicit specifications of domain knowledge are useful for new researchers who must learn the concepts and relations in the domain. In the scientific literature there are often synonyms such as reference point, reference vector and goal vector, they can be defined as identical individuals in the ontology to avoid confusion.

(4) \textit{To separate domain knowledge from the operational knowledge}.  In some sense developing an ontology is like  defining a set of data and their structure for other users or softwares to use. Different types of users, domain-independent applications and software agents use ontologies as input data. For instance, the PMOMH  ontology might serve an automatic text-mining or literature retrieval system to understand the meaning of certain concepts or words (e.g., to identify synonyms or subsumption of keywords).

(5) \textit{To analyze domain knowledge}. Usually an ontology is not the ultimate goal in itself. More useful information can be gained by analyzing the ontology. By building the PMOMH ontology we can easily analyze what kind of preferences have been integrated in what kind of MOMH, through what kind of integration. We can also query for methods that can deal with a specific kind of problem, or find potential combinations of MCDM and MOMH for future research. One possible use is to extract building blocks that can be used as operators within different kinds of MOMHs and design new PMOMHs.

There are various ontology applications in different fields, such as  Recommender Systems, e-Learning, e-Commerce, Semantic Interoperability, Bioinformatics, etc.  As far as we know, Evolutionary Computation (EC) ontology \cite{KaurChaudhary2015} and  diversity-oriented optimization ontology \cite{Basto-FernandesYevseyevaEmmerich2013} have been proposed recently, and their contributions were considered in the design of the PMOMH ontology. An ontology of preference-based multi-objective evolutionary algorithms (PMOEAs) was built and basic query examples were given \cite{li2017building}. In this paper we extend and improve the ontology, and provide more use cases to demonstrate the benefits of the provided ontology.

\section{Building the PMOMH Ontology}
\label{SectionOntologyBuilding}
We follow a bottom up (or inductive) approach to build the ontology, by first looking at the existing work (as Section \ref{SectionBackground} does) and then structuring it into an ontology. 
An OWL ontology consists of classes, properties (including object properties and data properties) and individuals. For the sake of readability, in the remaining of this paper, class names are written in a font \textit{like this}, object property names are in a font \texttt{like this}, data property names are in a font \textsf{like this}, individual names are in a font  \textsl{like this}.

A class describes a group of concepts with the same properties in the domain. 
For example, \textit{MOMH} is the class of multi-objective metaheuristics. A class can have subclasses that represent concepts more specific than the superclass. 
For example, \textit{PMOMH} is a subclass of \textit{MOMH} which uses preferences to focus the search on preferred parts of the PF. Subclasses inherit all the properties of superclasses. An object property is a binary relation to relate classes or  individuals. For instance, \texttt{canSolve} is an  object property that can relate \textit{MOMH} and \textit{MOP}, which indicates the capability of one specific MOMH with regard to solving one specific problem. A data property relates classes or individuals with a designed primitive data-type (e.g., integer, boolean, etc). For example, \textsf {hasPublishingYear} is a data property of \textit {MOMH} with datatype ``\textbf{integer}". Individuals represent objects, or class instances in the domain of interest, for instance\textsl{ R-NSGA-II} \cite{DebSundar2006} is an individual of \textit{PMOMH}.

As suggested by Noy \cite{NoyMcGuinnessothers2001}, there is no one ``correct" way for developing ontologies. However, the guidelines provided in \cite{NoyMcGuinnessothers2001} represent a generalized set of steps, widely accepted in this area, which are adopted by us and presented in detail next.

\subsection{Determine the domain and scope of the ontology}
The ontology we built addresses  preference-based multi-objective metaheuristics, or PMOMHs. They utilize the preference information provided by the DM, a-priori or interactively, to guide the search towards the interesting parts of the PF instead of searching for the whole set.

According to \cite{SindhyaRuizMiettinen2011}, at least two possible ways of integrating EMO and MCDM can be identified: ``evolutionary algorithm in MCDM" and ``MCDM in EMO" approaches. The former follows the main procedure of MCDM method and utilizes evolutionary algorithms to get intermediate solutions (such as PIE  \cite{SindhyaRuizMiettinen2011}), the latter, ``MCDM in EMO", or ``MCDM in MOMH" more broadly, was the main scope in our study.

We collected related literature on this topic (see Table  \ref{TablePMOEAClassification}), analyzed it and derived the important features to form an ontology. In addition to concepts and knowledge representation from PMOMH itself, the ontology also intends to provide support for questions like ``What methods can deal with a specific type of problem?", ``What is the implementation language of that algorithm?", etc. Therefore, detailed knowledge related to MOP, as well as optimization software frameworks are also considered in the ontology.

\subsection{Consider using the existing ontologies}

As far as we know, Evolutionary Computation (EC) Ontology \cite{KaurChaudhary2015}, Diversity-Oriented Optimization Ontology \cite{Basto-FernandesYevseyevaEmmerich2013} and Semantic MCDM (SeMCDM) \cite{mahmoudi2009semantic} are related to the PMOMH ontology to some extent. Because they were designed with strong focus on specific domain knowledge operationalization, we can not reuse them directly, but some common concepts and vocabularies can still be adopted.



Since the PMOMH, EC, SeMCDM and Diversity-Oriented Optimization ontologies define and use a subset of common concepts, this overlapping knowledge was taken into account in the PMOMH ontology design in two ways. The first way is to adopt the same naming for identical concepts, such as  \texttt{hasAuthor}, \texttt{hasPublishingYear}, \texttt{canSolve}, \textit{UtilityFunction}. The other way is to specify the relations between them using OWL relations: \texttt{owl:sameAs}, \texttt{ owl:equivalentProperty} and \texttt{ owl:equivalentClass}, to map identical individuals, properties and classes  between ontologies, respectively. For example,  \textit{SetQualityIndicator} of the PMOMH ontology is \texttt{owl:equivalentClass} to \textit{Indicator} of Diversity-Oriented Optimization Ontology.

\subsection{Enumerate important terms in the ontology}

It is important to make a comprehensive list of terms related to the PMOMH domain, without thinking of the overlap between some concepts. They are modeled as classes and properties in the next three sections.

\subsection{Define the classes and the class hierarchy}
%
%

The classes and class hierarchy of our ontology include the following, all of them are extendable if needed:

\begin{figure*}[!t]
\centering
\includegraphics [width=3in]{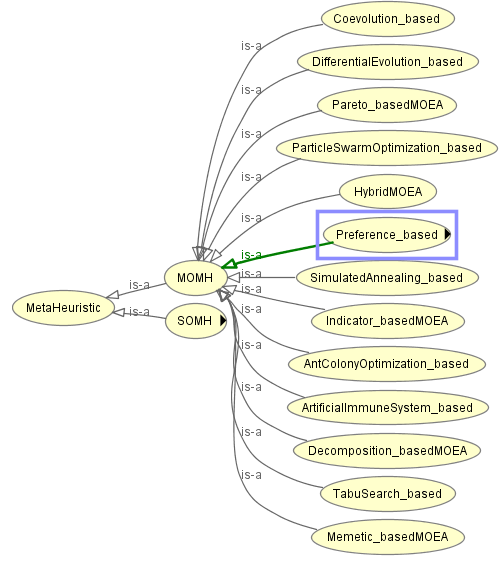}

\caption{Class hierarchy of Mataheuristic}
\label{Metaheuristic}
\end{figure*}

\textit{MetaHeuristic} class (Fig. \ref{Metaheuristic}) has \textit{ MOMH} (multi-objective metaheuristics) and \textit{SOMH} (single-objective metaheuristics) as subclasses. \textit{Preference\_based} is the class of PMOMH. It is a subclass of \textit{ MOMH}.  Other subclasses of \textit{ MOMH} include\textit{ Pareto\_basedMOEA},  \textit{Indicator\_basedMOEA}, \textit{ Decomposition\_basedMOEA}, \textit{ ParticleSwarmOptimization\_based},   \textit{ Coevolution\_based}, \textit{Memetic\_basedMOEA}, \textit{ HybridMOEA},  \textit{ SimulatedAnnealing\_based}, \textit{ DifferentialEvolution\_based}, \textit{ TabuSearch\_based}, \textit{ AntColonyOptimization\_based},  \textit{ ArtificialImmuneSystem\_based} . Note that there can be overlaps between classes, one individual can belong to several classes such as \textit{hybridMOEA} individuals may also belong to, for example \textit{ParticleSwarmOptimization\_based} or \textit{ ArtificialImmuneSystem\_based}. In order to keep the design transparent we opted for a flat structure. If a combination is not allowed, it can be added as a constraint using the ``disjoint with" property. One important property to link PMOMH and the other MOMH subclasses is \texttt{hasSearchAlgorithm} (can be found in Table \ref{TableObjectProperty}), which indicates the searching method used by the PMOMH.

\textit{MOP} is the class of multi-objective optimization problem. Subclasses include \textit{Academic\_Problem} and \textit{Realworld\_Problem}. \textit{Academic\_Problem} has subclasses \textit{DTLZ},  \textit{Knapsack},  \textit{WFG},  \textit{ZDT}, \textit{UF}.

\textit{InteractionTime} class indicates the moment when the DM interacts with the optimization process, where \textsl{ a-priori},\textsl{ a-posteriori} and\textsl{ progressive} are individuals.

\begin{figure*}[!t]
\centering
\includegraphics [width=4.5in]{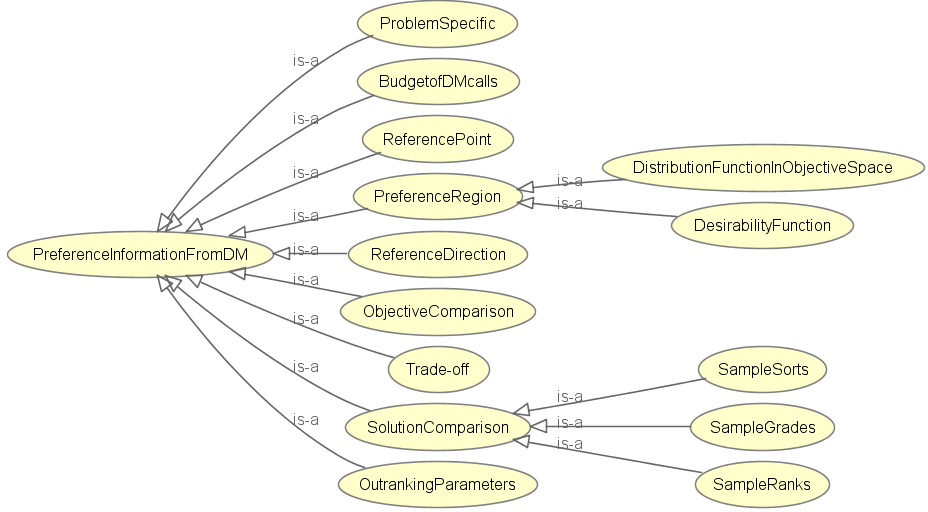}

\caption{Class hierarchy of PreferenceInformationFromDM}
\label{PreferenceInformationFromDM}
\end{figure*}

\textit{PreferenceInformationFromDM} class (Fig. \ref{PreferenceInformationFromDM}) refers to the information provided by the DM to express his/her preferences. Subclasses include \textit{BudgetofDMcalls}, \textit{SolutionComparison} (\textit{SampleGrades}, \textit{SampleRanks} and \textit{ SampleSorts} are subclasses),   \textit{ ReferencePoint},\textit{ PreferenceRegion} (\textit{DesirabilityFunction} and \textit{DistributionFunctionInObjectiveSpace} are subclasses), \textit{ReferenceDirection}, \textit{ Trade-off}, \textit{ObjectiveComparison}, \textit{OutrankingParameters}, \textit{ProblemSpecific}.

\begin{figure*}[!t]
\centering
\includegraphics [width=2.5in]{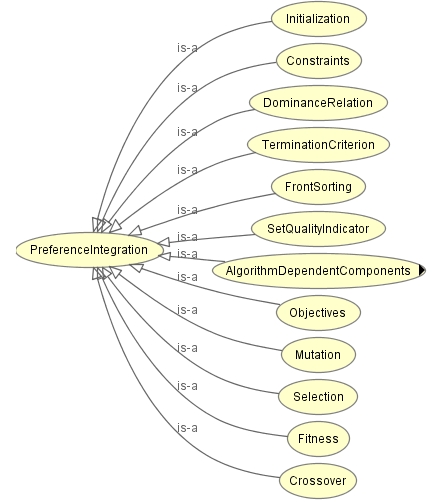}

\caption{Class hierarchy of PreferenceIntegration}
\label{PreferenceIntegration}
\end{figure*}

\textit{PreferenceIntegration} class (Fig. \ref{PreferenceIntegration}) defines how the preference information is integrated in the search method, i.e., what is modified in the searching algorithm to accommodate preferences. Subclasses are the following: \textit{ AlgorithmDependentComponents} (such as crowding distance in NSGA-II, weights in MOEA/D), \textit{DominanceRelation}, \textit{Objectives},\textit{ SetQualityIndicator},\textit{ Constraints},\textit{  TerminationCriterion}, \textit{ Selection}, \textit{Crossover},\textit{ Mutation},\textit{  Fitness},\textit{  FrontSorting},\textit{  Initialization}.

\begin{figure*}[!t]
\centering
\includegraphics [width=4in]{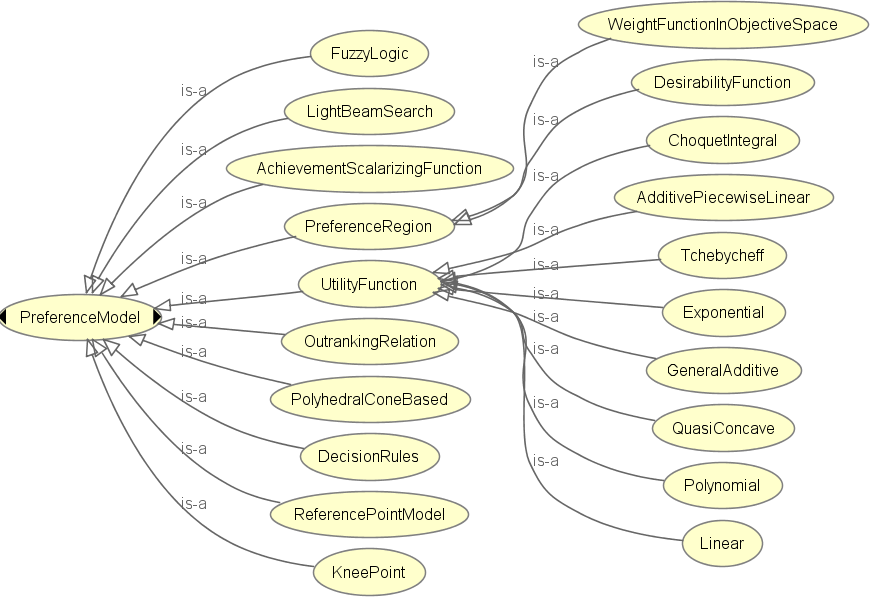}

\caption{Class hierarchy of PreferenceModel}
\label{PreferenceModel}
\end{figure*}

\textit{PreferenceModel} class (Fig. \ref{PreferenceModel}) specifies the preference model used in the PMOMH. It is strongly related to \textit{ PreferenceInformationFromDM}, but it focuses on the internal model utilized by the algorithm, which DM does not care or know. \textit{PreferenceModel}  subclasses include  \textit{AchievementScalarizingFunction}, \textit{FuzzyLogic},\textit{ DecisionRules},\textit{ OutrankingRelation},\textit{ PolyhedralConeBased}, \textit{UtilityFunction}  (which has\textit{ Linear}, \textit{AdditivePiecewiseLinear},\textit{ GeneralAdditive},\textit{ ChoquetIntegral},\textit{ Polynomial}, \textit{Quasiconcave}, \textit{Tchebycheff}, \textit{Exponential} as subclasses), \textit{ PreferenceRegion} (\textit{ DesirabilityFunction} and \textit{ WeightFunctionInObjectiveSpace} are subclasses), \textit{LightBeamSearch}, \textit{ReferencePointModel}, \textit{KneePoint}.

%

\textit{LearningMethod}  class is used to indicate the learning method of some PMOMHs (usually interactive PMOMHs) for predicting the DM's preferences. \textit{LearningMethod} subclasses include \textit{OrdinalRegression}, \textit{LinearProgramming}, \textit{QuadraticProgramming}, \textit{SupportVectorMachine}, \textit{ NeuralNetwork},  \textit{ SuperisedLearning}.

\textit{ResultType} class defines the type of the result, which can be classified as \textit{OneSolution} and \textit{SetOfSolutions}. \textsl{BiasedDistribution} and \textsl{PartialRegion} are individuals of \textit{SetOfSolutions}.

\textit{Researcher} class specifies the authors of the PMOMH papers.

\textit{ImplementationLibrary} class indicates the library or framework used by metaheuristics, such as \textsl{jMetal}, \textsl{KanGAL}, \textsl{PISA}, \textsl{MOEAFramework}. \textit{ProgrammingLanguage} class specifies the language used for implementation.

\subsection{Define the relations between classes (Object Properties)}

\begin{table}
\centering
\caption{Object Properties}
\label{TableObjectProperty}
\begin{tabular}{|c|c|c|}

\hline

Object Property & Domain & Range\\

\hline

 \texttt{hasResultType} & \textit{PMOMH} & \textit{ResultType} \\  
 \texttt{hasPreferenceModel} & \textit{PMOMH} &\textit{PreferenceModel} \\ 
 \texttt{canSolve} & \textit{MetaHeuristic} &\textit{MOP} \\ 
 \texttt{hasSearchAlgorithm} & \textit{PMOMH} &\textit{MOMH} \\ 
 \texttt{hasInteractionTime} & \textit{PMOMH} &\textit{InteractionTime} \\ 
 \texttt{hasAuthor} & \textit{MetaHeuristic} &\textit{Researcher} \\ 
 \texttt{hasPreferenceInformationFromDM} & \textit{PMOMH} &\textit{PreferenceInformationFromDM} \\ 
 \texttt{hasPreferenceIntegration} & \textit{PMOMH} &\textit{PreferenceIntegration} \\ 
 \texttt{hasLearningMethod} & \textit{PMOMH} &\textit{LearningMethod} \\
 \texttt{isInteractiveVersionOf} & \textit{PMOMH} &\textit{PMOMH}\\ 
 \texttt{hasInteractiveVersion} & \textit{PMOMH} &\textit{PMOMH}\\ 
 \texttt{hasComparison} & \textit{MetaHeuristic} &\textit{MetaHeuristic}\\ 
 \texttt{isExtensionOf} & \textit{MetaHeuristic} &\textit{MetaHeuristic}\\ 
 \texttt{hasExtension} & \textit{MetaHeuristic} &\textit{MetaHeuristic}\\ 
 \texttt{useLibrary} & \textit{MetaHeuristic} &\textit{ImplementationLibrary}\\ 
 \texttt{useLanguage} & \textit{MetaHeuristic} &\textit{ProgrammingLanguage}\\ 

\hline

\end{tabular}
\end{table}

Object properties are binary relations on individuals (of classes), they determine how the classes are related to each other. 
For example, in the assertion ``\textsl{ R-NSGA-II}  \texttt{hasPreferenceInformationFromDM} \textsl{ReferencePoint}", \texttt{ hasPreferenceInformationFromDM} is an object property whose domain is \textit{PMOMH} and range is \textit{PreferenceInformationFromDM}, indicating this property is from class \textit{PMOMH} to class  \textit{PreferenceInformationFromDM}, not the other way around.  The main object properties in our ontology are listed in Table \ref{TableObjectProperty}. Note that one individual can be related to several individuals with the same object property,  such as `` \textsl{R-NSGA-II}   \texttt{hasPreferenceInformationFromDM}  \textsl{ReferencePoint}" and \textsl``{ R-NSGA-II}  \texttt{hasPreferenceInformationFromDM}  \textsl{weights}" hold at the same time.

\begin{figure*}[!t]
\centering
\includegraphics [width=5.7in]{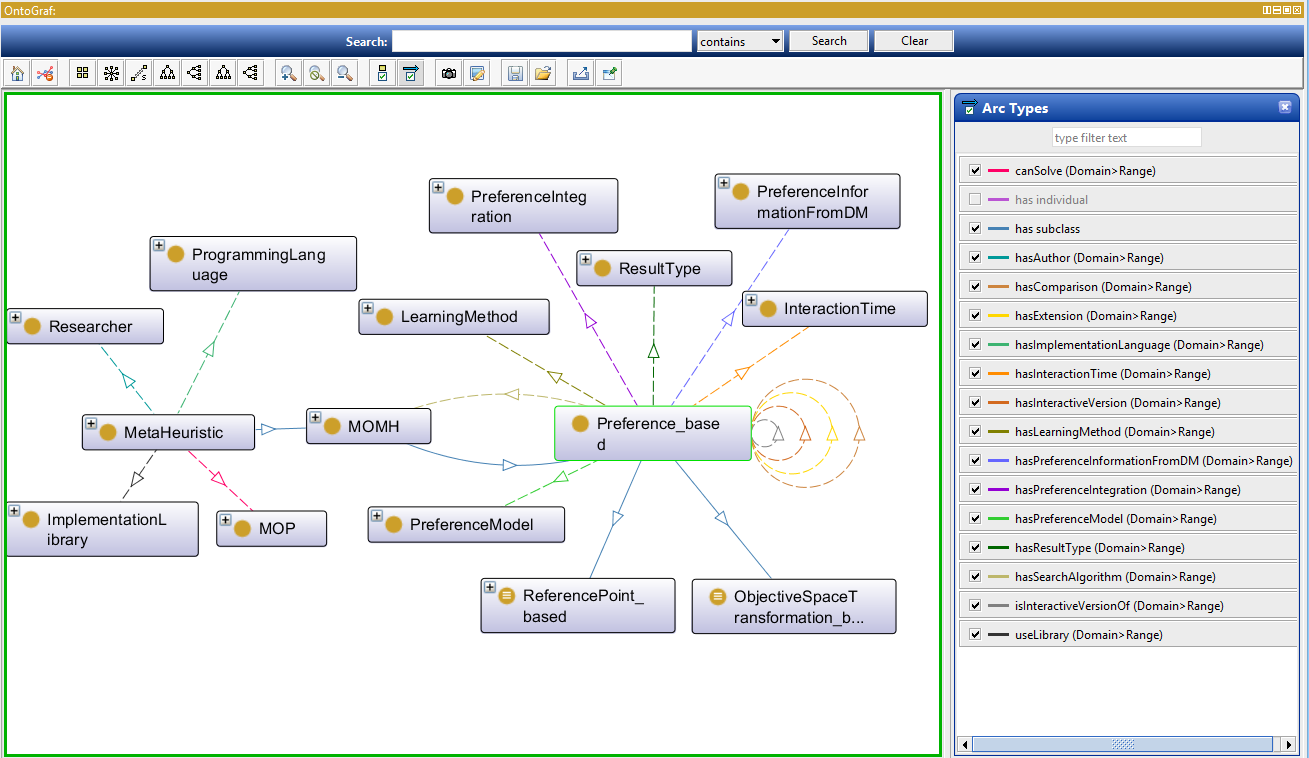}

\caption{Visualization with OntoGraf}
\label{Figure5}
\end{figure*}

\emph{OntoGraf}  \cite{a} is a (Prot\'eg\'e plugin) visualization tool that allows visual, interactive navigation of the relationships in OWL ontologies.  As Fig.\ref{Figure5}  shows, all the classes that are related to \textit{PMOMH} are rectangles and object properties are lines connecting the rectangles.

\subsection{Define the properties of classes (Data Properties)}
\begin{table}
\centering
\caption{Data Properties}
\label{TableDataProperty}
\begin{tabular}{|c|c|c|}

\hline

Data Property & Domain & Range\\

\hline

 \textsf{isContinuousProblem} & \textit{MOP} & \textbf{boolean} \\  
 \textsf{isDiscreteProblem} & \textit{MOP} & \textbf{boolean} \\
 \textsf{isMixedIntegerProblem} & \textit{MOP} & \textbf{boolean}\\ 
 \textsf{isManyObjectiveProblem} & \textit{MOP} & \textbf{boolean} \\ 
 \textsf{isMultimodalProblem} & \textit{MOP} & \textbf{boolean} \\ 
 \textsf{isNoisyProblem} & \textit{MOP} & \textbf{boolean} \\ 
 \textsf{hasExpensiveEvaluation} &  \textit{MOP} & \textbf{boolean} \\ 
 \textsf{hasNumberOfObjectives} & \textit{MOP} &\textbf{integer} \\ 
 \textsf{hasPublishingYear} & \textit{MetaHeuristic} &\textbf{integer}  \\
 \textsf{hasReference} & \textit{MetaHeuristic, MOP} &\textbf{string}  \\ 
  \textsf{hasCitationTimes} & \textit{MetaHeuristic} &\textbf{integer}  \\ 
 \textsf{hasMultipleRegionOfInterest} & \textit{PMOMH} &\textbf{boolean} \\ 
 \textsf{hasSpreadControl} & \textit{PMOMH} &\textbf{boolean}\\ 
 \textsf{preservesParetoDominance} & \textit{PMOMH} &\textbf{boolean} \\

\hline

\end{tabular}
\end{table}
Data properties link an individual to an XML Schema Datatype value or an RDF literal. In other words, they describe relationships between an individual and data values. For example, \textsf {hasPublishingYear} is a data property of \textsl{R-NSGA-II} with data type ''\textbf{integer}'', which indicates the year when \textsl{ R-NSGA-II} was published. The main data properties defined in our ontology are listed in Table \ref{TableDataProperty}.

There are three data properties that describe the characteristics of PMOMH: \textsf{hasMultipleRegionOfInterest} indicates whether the method offers DM the ability to obtain more than one ROI in one run; \textsf{hasSpreadControl} describes whether the method allows the DM to control the spread of the obtained ROI; \textsf{preservesParetoDominance} shows whether the method preserves the order induced by Pareto dominance. These are important properties of PMOMH, which are also examined in \cite{bechikh2015chapter}.

\subsection{Create Individuals}

Individuals of \textit{PMOMH} are algorithms named after the original paper or abbreviations of the method. We created every \textit{PMOMH} individual by reading the paper and answering the following questions (answers are used for property in the parentheses):

(1) What preference information should the DM provide? (\texttt{hasPreferenceInformationFromDM})

(2) When should s/he provide the preference information?  (\texttt{hasInteractionTime})

(3) What is the preference model of this algorithm?  (\texttt{hasPreferenceModel})

(4) Which MOMH is used in this method? (\texttt{hasSearchAlgorithm})

(5) How is the preference information integrated in the searching algorithm?  (\texttt{hasPreferenceIntegration})

(6) If there is a learning method in this algorithm, what is it?  (\texttt{hasLearningMethod})

(7) What type of result is obtained by the algorithm: one solution or set of solutions?  (\texttt{hasResultType})

(8) Who introduced this algorithm and when?  (\texttt{hasAuthor},  \textsf{hasPublishingYear})

(9) What problems are tested in the simulation experiments by this algorithm?  (\texttt{canSolve})

(10) Can the algorithm deal with multiple ROI in one run?  (\textsf{hasMultipleRegionOfInterest})

(11) Does the algorithm have spread control methodology if its result is part of the PF? (\textsf{hasSpreadControl})

(12) Is it compatible with Pareto dominance relation? (\textsf{preservesParetoDominance})

(13) What other PMOMHs  are compared in this paper?  (\texttt{hasComparison})

(14) Does the algorithm have an extension or interactive version (if it is of a-priori type)? (\texttt{hasExtension}, \texttt{hasInteractiveVersion})

(15) What library and programming language are used to implement this algorithm? (\texttt{useLibrary}, \texttt{useLanguage})

\begin{figure*}[!t]
\centering
\includegraphics [width=5.7in]{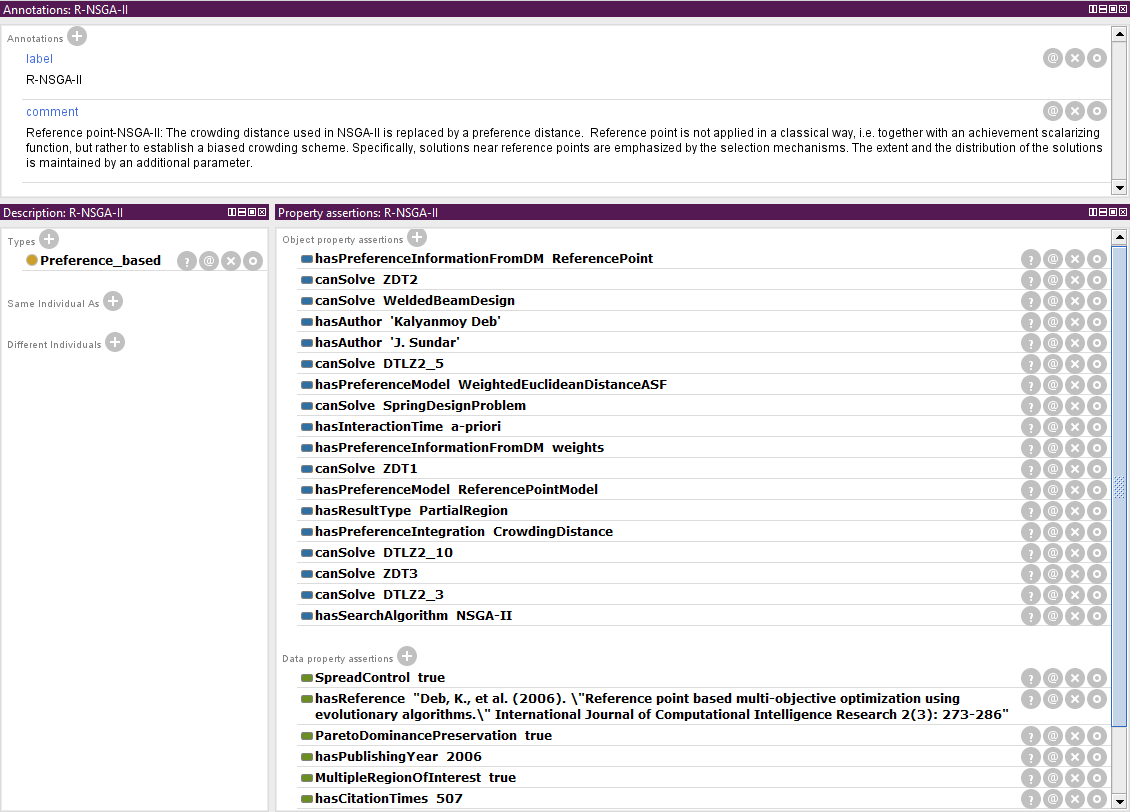}

\caption{R-NSGA-II individual}
\label{createIndividualR-NSGA-II}
\end{figure*}

Fig.\ref{createIndividualR-NSGA-II}  presents \textsl{R-NSGA-II}  \cite{DebSundar2006} as an example of  PMOMH individual.

Individuals of \textit{MOP}, another important class in the proposed PMOMH ontology, were created by describing their data properties as shown in Table \ref{TableDataProperty}.

%
Table \ref{Number of individuals} shows individual numbers of the most relevant classes in the current PMOMH ontology.

\begin{table}
\centering
\caption{Number of individuals}
\label{Number of individuals}
\begin{tabular}{|c|c|c|c|c|c|}

\hline

Class Name & \textit{PMOMH} & \textit{MOP} & \textit{ImplementationLibrary}&\textit{Researcher}& \textit{PreferenceModel} \\

\hline
Number & 86 & 158&15&166&42\\  

\hline

\end{tabular}
\end{table}

\subsection{Publish and reuse}
We have uploaded our ontology into the WebProt\'eg\'e (\url{http://webprotege.stanford.edu/#Edit:projectId=8c2f9a29-82ff-4f4b-9d56-676741213d66}).  WebProt\'eg\'e is an ontology development environment that makes it easy to create, upload, modify and share ontologies for collaborative viewing and editing.  Interested researchers and other professionals can view, comment, edit (after permission) and download the ontology for reuse and research.

\section{Using the PMOMH Ontology}
\label{SectionOntologyUsecases}
Since an ontology has been created, a lot of knowledge management tasks can be performed using the ontology together with other tools such as reasoners, DL Query and visualization tools.
\emph{Reasoners} can check the consistency of an ontology, perform subsumption checking (check if a concept is a subset of another concept) and infer relations that are not explicitly given by the user. \textit{DL Query} is another important feature of  Prot\'eg\'e, which can help to access, analyze and explore the knowledge domain described in the ontology.
Examples of reasoning and query are available in \cite{li2017building}. Based on these functions, we can use the PMOMH ontology for different purposes, which are given next.

\begin{figure*}[!t]
\centering
\includegraphics [width=4in]{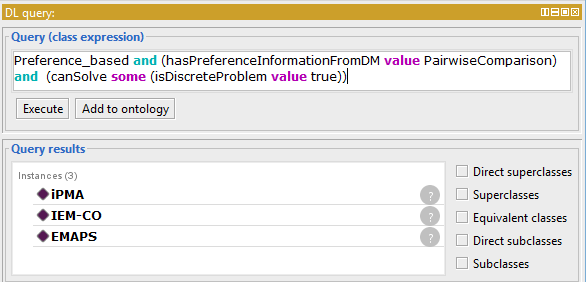}

\caption{Find appropriate algorithms for an application}
\label{Application}
\end{figure*}

\subsection{Finding the right method for an application}
Application to real world problems is the ultimate goal of developing various  PMOMHs. Different problems have diverse characteristics such as type of the decision variables used, e.g., continuous/discrete; or number of objectives, e.g., multi- or many-objectives;  or shape of PF, e.g., convex, concave, disconnected. PMOMH ontology can help to find methods that suit a specific problem.

For example, a satellite task scheduling problem requires to optimize the execution plan for the satellite in order to achieve certain objectives. It can be modeled as a multi-objective combinatorial optimization problem similar to the well-known knapsack problem. The objectives to consider are maximizing the total profit and minimizing the total cost. If the DM would like to use  pairwise comparison as preference articulation, then we can search the PMOMH ontology for a feasible algorithm for this problem. Fig. \ref{Application} shows that iPMA \cite{Jaszkiewicz2007}, IEM-CO \cite{PhelpsKoeksalan2003} and EMAPS  \cite{koksalan2007evolutionary} have pairwise comparison as preference articulation and solve discrete problems in the experiment, which are reasonable suggestions for our application problem. We do not assume that other PMOMHs can not solve the satellite scheduling problem, but recommend the indicated approaches since they are easy to apply with regard to the specified preference type and problem type.

\subsection{Bibliomatric Analysis}
\begin{figure*}[!t]
\centering
\includegraphics [width=3.5in]{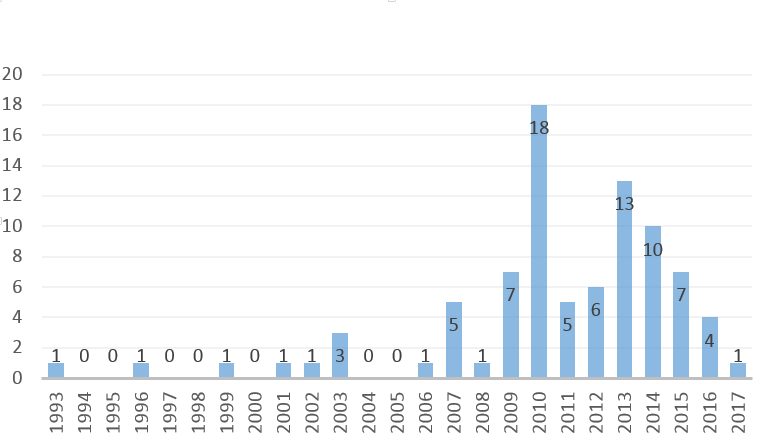}

\caption{Year distribution of PMOMH numbers}

\label{NumberByYear}
\end{figure*}

Bragge et al. conducted bibliomatric analysis on Multiple Criteria Decision Making/Multiattribute Utility Theory (MCDM/MAUT) \cite{bragge2010bibliometric}. The statistical analysis shows the change in the number of publications over time, top-20 lists of countries, authors, journals, presenting  a ``big picture" of MCDM/MAUT. Similar analysis can be conducted on PMOMH domain with the help of ontology. Fig. \ref{NumberByYear} shows the publishing year distribution of 86 PMOMH individuals.  Table \ref{MostCitedPMOMH} and Table \ref{ResearchersOfMostPMOMHNumber} reveal the top-5 most cited PMOMH individuals and top-5 researchers ranked by number of published PMOMH individuals regarding the 86 PMOMH individuals in the ontology.

\begin{table*}  
\begin{floatrow}  
\capbtabbox{  

\begin{tabular}{|c|c|c|}

\hline

Rank & PMOMH & CitedTimes  \\

\hline
 1 & \textsl{MOGA} \cite{FonsecaFlemingothers1993} & 4236\\  
 2 & \textsl{R-NSGA-II} \cite{DebSundar2006} &507 \\ 
3 & \textsl{NSGA-III} \cite{DebJain2014} &419 \\ 
4& \textsl{SIBEA} \cite{ZitzlerBrockhoffThiele2007} &325\\ 
5 &\textsl{G-MOEA} \cite{BrankeKauslerSchmeck2001}&228 \\ 
\hline

\end{tabular}
}
{

\caption{Top-5 most cited PMOMH individuals}
\label{MostCitedPMOMH}
}
\capbtabbox{ 

\begin{tabular}{|c|c|c|}

\hline

Rank & Researcher & Number  \\

\hline
1 & Kalyanmoy Deb  & 10\\  
2 & Heike Trautmann & 6 \\ 
2 & J\"urgen Branke &6 \\ 
2 & Licheng Jiao & 6\\ 
2 & Xiaodong Li& 6 \\ 
\hline

\end{tabular}
}
{
\caption{Top-5 researchers ranked by number of published PMOMH individuals}
\label{ResearchersOfMostPMOMHNumber}
}
\end{floatrow}
\end{table*}

%
%
%
%
%
%
%
%
%
%
%

\subsection{Reviewing and Research Assessment}
With the help of \emph{DL Query}, researchers can easily review and access the existing works. For example, they can find out whether a preference model or integration method  has been applied before, which papers compared their developed method(s) to one specified method (e.g., who compared their proposed method with  R-NSGA-II\cite{DebSundar2006}?) They can query for methods that used one specific benchmark problem (e.g., ZDT1) if they want to compare results on the same problem.  They can also query for relations of algorithms (e.g., extension and interactive version) or search for algorithms that use the same preference information (e.g., reference point).
\restoregeometry

\newgeometry{left=1.1cm}
\begin{table*}

\newcommand{\tabincell}[2]{\begin{tabular}{@{}#1@{}}#2\end{tabular}}
\caption{PMOMHs classified by searching algorithm and preference information }
\label{TablePMOEAClassification}
\small
\begin{tabular}{|c|c|c|c|c|c|}

\hline

\tabincell{c}{Searching algorithms\\Preferences} & Pareto-based MOEAs & Indicator-based MOEAs & other MOEAs & Alternative MOMHs\\

\hline

Desirability Function &\tabincell{c}{ \textsl{DF-NSGA-II}\cite{TrautmannMehnen2009}\\ \textsl{DI-EMOA}\cite{TrautmannWagnerBiermannEtAl2013}}&\tabincell{c}{\textsl{DHI}\cite{EmmerichDeutzYevseyeva2014}\\\textsl{DF-SMS-EMOA}\cite{WagnerTrautmann2010}} &&\textsl{DF-MOPSO}\cite{MostaghimTrautmannMersmann2010} \\  
\hline

Objective Comparison &\tabincell{c}{ \textsl{GPS-EA}\cite{TanKhorLeeEtAl2003}\\\textsl{FP-EMO}\cite{JinSendhoff2002}\\  \textsl{RIO-NSGA-II}\cite{RachmawatiSrinivasan2010}}&\textsl{iW-HypE}\cite{BrockhoffHamadiKaci2014} &\tabincell{c}{ \textsl{MQEA-PS}\cite{KimHanKimEtAl2012}} & \tabincell{c}{\textsl{MOPSO-PS}\cite{LeeKim2011}}\\
\hline
Preference Region/Distribution &\tabincell{c}{ \textsl{prTDEA}\cite{KarahanKoeksalan2010}\\ \textsl{W-NSGA-II}\cite{FriedrichKroegerNeumann2011}\\ \textsl{W-SPEA2}\cite{FriedrichKroegerNeumann2011}\\ \textsl{GF-NSGA-II}\cite{NarukawaSetoguchiTanigakiEtAl2015}}& \tabincell{c}{ \textsl{SIBEA}\cite{ZitzlerBrockhoffThiele2007}\\ \textsl{W-HypE}\cite{AugerBaderBrockhoffEtAl2009a}\\ \textsl{W-HypE\'}\cite{BrockhoffBaderThieleEtAl2013}\\ \textsl{TEHVI-EGO}\cite{yang2016preference}} & \tabincell{c}{\textsl{iPICEA-G}\cite{WangPurshouseFleming2013}\\\textsl{RVEA}\cite{ChengJinOlhoferEtAl2016}} &\tabincell{c}{\textsl{HMIA}\cite{jiao2010hybrid} }\\
\hline

	Reference Point& \tabincell{c}{\textsl{g-NSGA-II}\cite{MolinaSantanaHernandez-DiazEtAl2009}\\\textsl{GPS-EA}\cite{TanKhorLeeEtAl2003}\\\textsl{CP-NSGA-II}\cite{JaimesMontanoCoello2011}\\\textsl{r-NSGA-II}\cite{BenSaidBechikhGhedira2010}\\\textsl{R-NSGA-II}\cite{DebSundar2006}\\\textsl{MOGA}\cite{FonsecaFlemingothers1993}\\\textsl{2p-NSGA-II}\cite{Fei-yueYu-shiLi-pingEtAl2012}\\\textsl{SR-NSGA-II}\cite{filatovas2015synchronous}\\\textsl{ER-NSGA-II}\cite{siegmund2012finding}  }&  \tabincell{c}{\textsl{AS-EMOA}\cite{Trautmann2014}\\\textsl{PBEA}\cite{ThieleMiettinenKorhonenEtAl2009}\\\textsl{R2-EMOA}\cite{TrautmannWagnerBrockhoff2013}}&\tabincell{c}{\textsl{iPICEA-G}\cite{WangPurshouseFleming2013}\\\textsl{WASF-GA}\cite{RuizSaboridoLuque2015}\\\textsl{iWASF-GA}\cite{RuizLuqueMiettinenEtAl2015}\\\textsl{WZ-MOEA/D}\cite{rostami2015novel}\\\textsl{MOEA/D-PWA}\cite{qi2017reservoir}\\\textsl{R-MEAD}\cite{mohammadi2012reference}\\\textsl{R-MEAD2}\cite{MohammadiOmidvarLiEtAl2014}\\\textsl{NSGA-III}\cite{DebJain2014}} &\tabincell{c}{\textsl{RPSO-SS}\cite{AllmendingerLiBranke2008}\\\textsl{MDEPSO-RP}\cite{WickramasingheLi2009}\\\textsl{g-MOACO}\cite{chica2015interactive}}\\
\hline

Reference Direction& \tabincell{c}{\textsl{RD-NSGA-II}\cite{DebKumar2007}\\\textsl{LBS-NSGA-II}\cite{DebKumar2007a}\\\textsl{BCD-NSGA-II}\cite{BrankeDeb2005}}&\textsl{R2-EMOA}\cite{TrautmannWagnerBrockhoff2013}&\tabincell{c}{\textsl{iPICEA-G}\cite{WangPurshouseFleming2013a}\\\textsl{pMOEA/D}\cite{ma2016moea}}&\tabincell{c}{\textsl{MDEPSO-LBS}\cite{WickramasingheLi2009}\\\textsl{r-PMOA}\cite{liu2016r}\\\textsl{PRIMCSA}\cite{yang2010clone}\\\textsl{IPISA}\cite{liu2013preference}}\\

\hline

Trade-off&  \tabincell{c}{\textsl{G-MOEA}\cite{BrankeKauslerSchmeck2001}\\\textsl{pNSGA-II}\cite{ShuklaHirschSchmeck2010} }& \tabincell{c}{ \textsl{CHI-SMS-EMOA}\cite{ShuklaEmmerichDeutz2013}\\\textsl{CHI-EMOA}\cite{EmmerichDeutzKruisselbrinkEtAl2013}} &&\\
\hline

Pairwise Comparison&  \tabincell{c}{\textsl{NEMO-0}\cite{BrankeGrecoSlowinskiEtAl2015}\\\textsl{NEMO-I}\cite{BrankeGrecoSlowinskiEtAl2010}\\\textsl{NEMO-II}\cite{BrankeCorrenteGrecoEtAl2016}\\\textsl{IEM-CO}\cite{PhelpsKoeksalan2003}\\\textsl{DRSA-EMO-PCT}\cite{GrecoMatarazzoSlowinski2010}\\\textsl{IEA-SOL}\cite{krettek2009interactive}\\\textsl{INSPM}\cite{pedro2014inspm}\\\textsl{EMAPS}\cite{koksalan2007evolutionary}}& &\textsl{iPMA}\cite{Jaszkiewicz2007}&\\
\hline

Sample Ranks/Sorts &  \tabincell{c}{\textsl{PI-EMO-PC}\cite{SinhaKorhonenWalleniusEtAl2010}\\\textsl{PI-EMO-PC\'}\cite{SinhaKorhonenWalleniusEtAl2014}\\\textsl{iTDEA}\cite{KoeksalanKarahan2010}\\\textsl{BC-EMOA}\cite{BattitiPasserini2010}\\\textsl{PI-EMO-VF}\cite{DebSinhaKorhonenEtAl2010}\\\textsl{IEA-PP}\cite{gong2013evolutionary}\\\textsl{NN-DM-iTDEA}\cite{pedro2013decision}\\ \textsl{H-MCSGA}\cite{Cruz-ReyesFernandezGomezEtAl2014}\\
\textsl{DRSA-EMO}\cite{GrecoMatarazzoSlowinski2010}\\
\textsl{MCGA-ANN}\cite{todd1999directed}\\
\textsl{FFEA}\cite{greenwood1996fitness}} & \tabincell{c}{\textsl{iW-HypE}\cite{BrockhoffHamadiKaci2014}\\\textsl{SPAM}\cite{zitzler2010set}\\\textsl{I-SIBEA}\cite{chugh2015interactive}} & \tabincell{c}{\textsl{iMOEA/D}\cite{GongLiuZhangEtAl2011}\\\textsl{iEMOA-QCVF}\cite{FowlerGelKoeksalanEtAl2010}}& \\
\hline

Outranking Parameters & \tabincell{c}{\textsl{NOSGA}\cite{FernandezLopezBernalEtAl2010}\\\textsl{NOSGA-II}\cite{FernandezLopezLopezEtAl2011}}& &\tabincell{c}{\textsl{EvABOR-III}\cite{oliveira2013comparative}\\\textsl{HESA}\cite{oliveira2013hybrid}} &\\
\hline

Knee Point & \tabincell{c}{\textsl{KR-NSGA-II}\cite{BechikhBenSaidGhedira2010}\\\textsl{TKR-NSGA-II}\cite{BechikhSaidGhedira2011}}& & &\\

\hline

\end{tabular}
\end{table*}

\restoregeometry
\newgeometry{bottom=3.7cm}
\subsection{Classification}
There are various criteria to classify the PMOMHs, e.g., interaction time, preference information, search algorithm, to name a few.  Prot\'eg\'e can add the results of a query to new classes, thus making classification easy and flexible. 
For example, we can query for PMOMHs that use reference point as preference information and add the results to a new class \textit{ReferencePoint-based PMOMH}, which is a subclass of  \textit{PMOMH}. Any PMOMH to be added in the future that uses reference point will be automatically sorted into this subclass. 
Table \ref{TablePMOEAClassification}  shows one classification of PMOMHs based on  preference information and searching algorithms. 
%

\subsection{Identification of promising research areas}
Blank cells in the Table \ref{TablePMOEAClassification}  may be identified as future  research topics, to integrate preference articulation with MOMHs that such combination has not been considered before. Integration method and preference model can also be queried to find potential combinations. We can also notice the following:
\begin{itemize}
\item Not all the MCDM methods are included in this ontology. Classification of objective functions, for example, is one category of interactive MCDM methods \cite{MiettinenRuizWierzbicki2008}. At each interaction phase, the DM is shown the current Pareto optimal solution and asked to classify the objectives into several categories, i.e., whose values should be improved (till some desired aspiration level), whose values can be impaired (till some upper bound), whose values are temporarily allowed to change freely. This kind of MCDM method has not been integrated in MOMH, which can be a topic to consider in the future.

\item Some PMOMH individuals have interactive version (such as \textsl{W-HypE} \cite{AugerBaderBrockhoffEtAl2009a} and \textsl{iW-HypE} \cite{BrockhoffHamadiKaci2014}) while some do not. In theory all a-priori methods can be converted into a progressive method if the preference information changes during the search. By interaction with the algorithm, the DM can better understand the problem and give more precise preference information. For example, a preference region can be used at early stages of the optimization to focus the search within this region. After interaction  the DM may change the preference articulation to a precise reference point. New algorithms are needed to handle combined preference articulation.

\item Some PMOMH individuals use a new dominance relation to integrate preferences, such as g-dominance \cite{MolinaSantanaHernandez-DiazEtAl2009}, r-dominance \cite{BenSaidBechikhGhedira2010}. There are 13 \textit{DominanceRelation} individuals in the PMOMH ontology, they can be coupled with any other MOMHs without changing the main structure of the algorithm, thus new algorithms can be designed.

\item Some PMOMH individuals were proposed based on the same preference information, for example reference point. A performance metric is needed to compare different algorithms with the same preference articulation. Related works can be found in \cite{mohammadi2013new,ojalehto2016towards}, which are all related to reference point-based approaches, other preference models should also be considered.

\end{itemize}
\subsection{Maintenance and Extension }
The PMOMH ontology offers a structure, and is open for new classes/individuals to be added with the development of MCDM and MOMH fields.  In  WebProt\'eg\'e\, users can easily edit (e.g., revise properties, add comments), create (new individuals/classes/properties), delete (wrong/redundant  individuals/classes/properties) and download the ontology (OWL file). For example, people may reckon that \textit{Decomposition\_basedMOEA} should be named \textit{Aggregation\_basedMOEA} as in \cite{wang2016diversity}, this can be added as a comment to the class. With the discussion and collaboration of researchers from the related domains, the ontology will be more trustworthy and robust. The value of the proposed ontology increases with the progressive accumulation of more information. MOMH researchers can help to extend the MOMH classes,  MCDM researchers can assist with the preference information/model classes, MOP class can also be extended by adding more practical engineering applications. In a nutshell, PMOMH ontology helps to bring together researchers of different fields as well as practitioners who use the ontology. It provides a knowledge sharing platform for both academic research and  real-world application.

\section{Conclusion}
\label{SectionConclusion}
Preference-based multi-objective metaheuristics, or PMOMH, which combine MOMH and MCDM fields, are addressed in this paper. We proposed a novel method to systematize and manage the current knowledge in this field, by means of ontology design and engineering. For the first time, an ontology was designed and used in this field, showing high benefits and advantages for knowledge management in the MOMH and MCDM domains. After providing an overview of PMOMHs, we presented the process for building the PMOMH ontology using Prot\'eg\'e, introduced typical use cases, including application-method matching, bibliomatric analysis, research assessment and new research opportunities discovery.  The result of this article is representing the knowledge in a formal, computer-readable, way. This will allow other researchers to state specific queries. Moreover, we view the ontology as an evolving system and want to encourage integration of future research results.

It is believed that, the more attention and information PMOMH ontology attracts, the higher its value will be, not only for the research communities of MOMH and MCDM, but also for practitioners who are using the PMOMHs. Therefore, reuse and extension of the proposed PMOMH ontology are welcome to help its growth. In the future, new MCDM, MOMH and PMOMH individuals will be added to the ontology when they are found important. Meanwhile, benchmarks of practical applications, implementation source codes of PMOMHs will be collected and considered.

\paragraph{Acknowledgement:} Longmei Li acknowledges financial support from China Scholarship Council (CSC). Heike Trautmann and Michael Emmerich acknowledge support by the European Research Center for Information Systems (ERCIS).

\section*{References}

\bibliography{mybibfile}
\end{document}